\documentclass[12pt]{article}

\usepackage{arxiv}
\usepackage{amsmath, amssymb, amsthm}
\usepackage{bm}
\usepackage{graphicx}
\usepackage{booktabs}
\usepackage{hyperref}
\usepackage{tikz}
\usepackage{pgfplots}
\pgfplotsset{compat=1.18}

\hypersetup{
  colorlinks=true,
  linkcolor=blue,
  citecolor=blue,
  urlcolor=blue,
  pdftitle={Self-Attention as Distributional Projection: A Unified Interpretation of Transformer Architecture},
  pdfauthor={Nihal Mehta}
}

\title{Self-Attention as Distributional Projection: A Unified Interpretation of Transformer Architecture}

\author{
Nihal Mehta \\
Independent Researcher \\
Ph.D.\ (1990), University of Pennsylvania \\
\small \href{https://orcid.org/0009-0008-8012-3890}{ORCID: 0009-0008-8012-3890}
}

\date{\today}

\theoremstyle{plain}
\newtheorem{proposition}{Proposition}

\begin{document}
\maketitle

\begin{abstract}
This paper presents a mathematical interpretation of self-attention by connecting it to distributional semantics principles. We show that self-attention emerges from projecting corpus-level co-occurrence statistics into sequence context. Starting from the co-occurrence matrix underlying GloVe embeddings, we demonstrate how the 
projection naturally captures contextual influence, with the query-key-value mechanism arising as the natural asymmetric extension for modeling directional relationships. 
Positional encodings and multi-head attention then follow as structured refinements of this same projection principle. Our analysis demonstrates that the Transformer architecture's particular algebraic form follows from these projection principles rather than being an arbitrary design choice.
\end{abstract}
\vspace{0.5em}
\section{Introduction}
Transformer models rely on self-attention layers to construct contextual token representations \cite{vaswani2017attention}.
These layers are typically introduced procedurally: given learned projections $W_Q$, $W_K$, $W_V$, a sequence of token embeddings $H$ is mapped to queries, keys, and values; pairwise dot products between queries and keys are normalized by row-wise softmax to produce attention weights. Despite their empirical success, this mechanism is rarely analyzed from mathematical principles, leaving open why this particular algebraic form so effectively models contextual interaction.

The standard presentation borrows terminology from information retrieval---queries search for matching keys to retrieve values---but this analogy obscures rather than illuminates. The equations $\mathrm{softmax}(QK^\top/\sqrt{d_k})V$ contain no search procedure; the $Q$, $K$, $V$ matrices are simply learned projections whose relationship to ``queries'' and ``keys'' is purely metaphorical. This borrowed language has left the architectural choices appearing arbitrary and the mechanism's effectiveness mysterious.

This paper presents a mathematical analysis that addresses this gap by extending the distributional semantics foundation of GloVe \cite{pennington2014glove}. We begin with the symmetric co-occurrence operator $S\in\mathbb{R}^{n\times n}$ that underlies GloVe embeddings, projecting it into sequence-specific contexts via selector matrix $Q\in\mathbb{R}^{R\times n}$ to capture how words interact in context. This bridges the gap between classical distributional semantics and modern neural architectures, offering a principled explanation for the architectural choices in Transformers.

This work provides a mathematical clarification of the self-attention mechanism, showing that the Transformer architecture emerges naturally from extending distributional semantics principles to handle the core challenges of sequence modeling.

This framework reveals three key insights that connect distributional semantics to Transformer architecture:

\begin{itemize}
    \item \textbf{Projection Characterization:} The projection of global co-occurrence statistics into local sequence context is uniquely characterized by natural mathematical properties, providing a principled foundation for understanding the algebraic structure of self-attention.
    
    \item \textbf{Asymmetric Extension:} The query-key split emerges as the natural extension when generalizing the symmetric projection to model directional linguistic relationships, explaining why Transformers use separate transformations rather than a single symmetric operator.
    
    \item \textbf{Unified Framework:} Positional encodings and multi-head attention arise as structured refinements of the same projection principle, revealing how the complete architecture follows from systematically addressing the requirements of sequence modeling.
\end{itemize}

This progression—from symmetric projection through asymmetric generalization to multi-head factorization—demonstrates that the Transformer architecture is not merely an empirically successful design, but rather the mathematical consequence of systematically extending distributional semantics to handle the core challenges of sequence modeling: context specificity, directional relationships, and positional order.

Our analysis proceeds systematically: we begin with the symmetric co-occurrence projection $M = QSQ^\top$ (Section~\ref{sec:projection}), extend it to vocabulary-level prediction (Section~\ref{sec:prediction}), incorporate positional sensitivity (Section~\ref{sec:positional}), then introduce directional attention through learned transformations (Section~\ref{sec:directional}), and finally connect to the full Transformer architecture including multi-head attention (Section~\ref{sec:multihead}). Throughout, we emphasize the mathematical intuition behind each component, with illustrative examples included solely to clarify the analysis.

% ------------------------------------------------------

\section{Projection of Corpus-Level Statistics into Sequence Context}
\label{sec:projection}

The standard presentation of self-attention begins with learned projections $W_Q, W_K, W_V$ applied to token embeddings. This starting point, however, already assumes the architectural form we wish to understand. We instead begin with the most fundamental statistical regularities in language: distributional structure as captured through co-occurrence statistics.

\subsection{Global co-occurrence as a representation operator}

Let the corpus contain $n$ distinct token types. Define the global co-occurrence operator $S \in \mathbb{R}^{n\times n}$, where each entry $S_{ij}$ measures the empirical association between tokens $i$ and $j$. Each row of $S$ represents the empirical neighborhood of a token---the distribution of other tokens that tend to appear near it. Multiplying a one-hot basis vector $e_i$ by $S$ yields a vector $e_i S$ whose components indicate the relative frequencies of words co-occurring with token $i$. Thus, $S$ defines a linear map from token identity to its observed contextual distribution.

\subsection{Restricting to the input subspace}

Consider an input sequence of $R$ tokens. Define a selection matrix $Q \in \mathbb{R}^{R\times n}$ whose rows identify the tokens that appear in the input. The local projection
\[
M=QSQ^{\top}\in\mathbb{R}^{R\times R},
\]
restricts the global operator $S$ to the subspace spanned by the input tokens. 
Left-multiplying by $Q$ retains only associations that originate from input tokens; 
right-multiplying by $Q^{\top}$ retains only associations that terminate on input tokens. 

The resulting matrix $M$ extracts from the global co-occurrence statistics the specific tokens that are contextually related to the input sequence while filtering out tokens that don't align with the emergent semantic context. \textbf{Formally, this projection extracts the subnetwork of relationships most relevant to the current sequence}, amplifying associations that are mutually reinforced by the input tokens and suppressing those that are contextually incompatible:
\[
M_{rs} = q_r S q_s^{\top},
\]
which measures the compatibility between tokens $r$ and $s$ according to the global distribution. Each row of $M$ therefore defines how one token's representation should be adjusted in light of the others that co-occur with it in data.

\subsection{Algebraic Analysis of Contextual Weighting}

The contextual weighting mechanism is defined as follows:
\begin{itemize}
    \item $q_r \in \mathbb{R}^n$ is the $r$-th row of $Q$. This row vector is a one-hot vector selecting the token at position $r$.
    \item $M_{rs} = q_r S q_s^\top$ be the contextual weight between tokens $r$ and $s$
\end{itemize}

The matrix $M$ encodes how much tokens agree about the semantic context. The amplification and suppression properties emerge from the co-occurrence patterns in $M$:

\textbf{Amplification} occurs when:
\begin{itemize}
    \item Multiple tokens have strong mutual agreement ($M_{rs} \gg 0$) due to high co-occurrence between the tokens
    \item These tokens share common associations with the same vocabulary items
    \item Result: Constructive interference in the contextual representation
\end{itemize}

\textbf{Suppression} occurs when:
\begin{itemize}
    \item Tokens have weak contextual agreement ($M_{rs} \approx 0$) due to low co-occurrence
    \item Vocabulary associations are isolated to single or few tokens
    \item Result: Dilution and weak evidence due to lack of reinforcement
\end{itemize}

This algebraic structure automatically amplifies semantically coherent associations while suppressing contextually isolated ones through the agreement patterns encoded in $M$.

\subsection{Row Normalization and Emergent Attention}

The matrix $M=QSQ^{\top}$ contains raw compatibility scores that require normalization to function as proper attention weights. We apply row-wise normalization to convert these scores into probability distributions:

\[
\mathrm{Norm}(M)_{rs} = \frac{M_{rs}}{\sum_{t=1}^R M_{rt}}
\]

This normalization serves two crucial purposes:
\begin{enumerate}
\item It converts the compatibility scores into probability-like distributions over contextual partners
\item It ensures numerical stability by controlling the magnitude of the scores
\end{enumerate}

The normalized matrix $\mathrm{Norm}(M)$ already functions as a primitive attention mechanism: each token redistributes its representation according to corpus-consistent co-occurrence weights. This indicates that the core attention mechanism emerges naturally from normalized projections of corpus statistics, with the learned scaling in Transformers serving as a refinement of this basic principle.

\subsection{Illustrative Examples}
\label{sec:projection_examples}

The role of the projection $M = QSQ^{\top}$ and its normalized form $\mathrm{Norm}(M)$ is illustrated through contrasting examples that use the same global co-occurrence matrix $S$. These examples show how the mechanism automatically performs contextual disambiguation, amplifying mutually reinforced associations while suppressing incompatible ones.

\paragraph{Example 1: Geographical context ("river bank flooded")}

Consider a vocabulary with clear semantic groupings:
\[
V=\{\text{river},\ \text{bank},\ \text{loan},\ \text{money},\ 
\text{flooded},\ \text{shore},\ \text{rely}\},
\]
and co-occurrence matrix:
\[
S =
\begin{array}{c|rrrrrrr}
   & \text{river} & \text{bank} & \text{loan} & \text{money} &
     \text{flooded} & \text{shore} & \text{rely} \\ \hline
\text{river}   & 0 & 4 & 0 & 0 & 5 & 6 & 0 \\
\text{bank}    & 4 & 0 & 6 & 5 & 3 & 5 & 4 \\
\text{loan}    & 0 & 6 & 0 & 7 & 0 & 0 & 2 \\
\text{money}   & 0 & 5 & 7 & 0 & 0 & 0 & 3 \\
\text{flooded} & 5 & 3 & 0 & 0 & 0 & 4 & 0 \\
\text{shore}   & 6 & 5 & 0 & 0 & 4 & 0 & 0 \\
\text{rely}    & 0 & 4 & 2 & 3 & 0 & 0 & 0
\end{array}
\]

For the input sequence ``river bank flooded,'' the selector matrix $Q\in\mathbb{R}^{3\times 7}$ has one row per token:
\[
Q =
\begin{bmatrix}
1 & 0 & 0 & 0 & 0 & 0 & 0 \\   % river
0 & 1 & 0 & 0 & 0 & 0 & 0 \\   % bank
0 & 0 & 0 & 0 & 1 & 0 & 0      % flooded
\end{bmatrix}.
\]

\paragraph{Row selection via $QS$.}
Left-multiplying by $Q$ selects the rows of $S$ corresponding to our input tokens:
\[
QS =
\begin{bmatrix}
1 & 0 & 0 & 0 & 0 & 0 & 0 \\
0 & 1 & 0 & 0 & 0 & 0 & 0 \\
0 & 0 & 0 & 0 & 1 & 0 & 0
\end{bmatrix}
\begin{bmatrix}
0 & 4 & 0 & 0 & 5 & 6 & 0 \\
4 & 0 & 6 & 5 & 3 & 5 & 4 \\
0 & 6 & 0 & 7 & 0 & 0 & 0 \\
0 & 5 & 7 & 0 & 0 & 0 & 0 \\
5 & 3 & 0 & 0 & 0 & 4 & 0 \\
6 & 5 & 0 & 0 & 4 & 0 & 0 \\
0 & 4 & 2 & 3 & 0 & 0 & 0
\end{bmatrix}
=
\begin{bmatrix}
0 & 4 & 0 & 0 & 5 & 6 & 0 \\
4 & 0 & 6 & 5 & 3 & 5 & 4 \\
5 & 3 & 0 & 0 & 0 & 4 & 0
\end{bmatrix}.
\]

This extracts the global association profiles for \texttt{river}, \texttt{bank}, and \texttt{flooded}. At this stage, \texttt{bank} shows connections to both senses:
- Geographical: \texttt{river}=4, \texttt{flooded}=3, \texttt{shore}=5  
- Financial: \texttt{loan}=6, \texttt{money}=5, \texttt{rely}=4

\paragraph{Column selection via $(QS)Q^\top$.}
Right-multiplying by $Q^\top$ selects only columns corresponding to input tokens:
\[
M = (QS)Q^\top =
\begin{bmatrix}
0 & 4 & 0 & 0 & 5 & 6 & 0 \\
4 & 0 & 6 & 5 & 3 & 5 & 4 \\
5 & 3 & 0 & 0 & 0 & 4 & 0
\end{bmatrix}
\begin{bmatrix}
1 & 0 & 0 \\
0 & 1 & 0 \\
0 & 0 & 0 \\
0 & 0 & 0 \\
0 & 0 & 1 \\
0 & 0 & 0 \\
0 & 0 & 0
\end{bmatrix}
=
\begin{bmatrix}
0 & 4 & 5 \\
4 & 0 & 3 \\
5 & 3 & 0
\end{bmatrix}.
\]

The matrix $M$ describes the induced relations among the three input tokens. The normalized form demonstrates the contextual disambiguation:

\[
\mathrm{Norm}(M)=
\begin{bmatrix}
0.00 & 0.44 & 0.56 \\
0.57 & 0.00 & 0.43 \\
0.62 & 0.38 & 0.00
\end{bmatrix}.
\]

\textbf{This illustrates the disambiguation mechanism:}

\begin{itemize}
\item \texttt{river} associates with: \texttt{bank} (44\%) and \texttt{flooded} (56\%)
\item \texttt{bank} associates with: \texttt{river} (57\%) and \texttt{flooded} (43\%)  
\item \texttt{flooded} associates with: \texttt{river} (62\%) and \texttt{bank} (38\%)
\end{itemize}

The projection has resolved the ambiguity of \texttt{bank} toward the geographical sense because:
\begin{enumerate}
\item \texttt{river} and \texttt{bank} share strong connections to \texttt{shore} (6 and 5 respectively)
\item \texttt{river} and \texttt{flooded} share strong connections (5)
\item The financial connections of \texttt{bank} find no support from the other input tokens
\end{enumerate}

This example demonstrates the conceptual operation of the projection: each token's attention is distributed among other tokens based on their co-occurrence relationships, with the pattern reflecting the contextual associations in the input sequence.

\paragraph{Example 2: Financial context ("bank loan")}

For the input sequence ``bank loan,'' the selector matrix $Q\in\mathbb{R}^{2\times 7}$ has one row per token:
\[
Q =
\begin{bmatrix}
0 & 1 & 0 & 0 & 0 & 0 & 0 \\   % bank
0 & 0 & 1 & 0 & 0 & 0 & 0      % loan
\end{bmatrix}.
\]

\paragraph{Row selection via $QS$.}
Left-multiplying by $Q$ selects the rows of $S$ corresponding to our input tokens:
\[
QS =
\begin{bmatrix}
0 & 1 & 0 & 0 & 0 & 0 & 0 \\
0 & 0 & 1 & 0 & 0 & 0 & 0
\end{bmatrix}
\begin{bmatrix}
0 & 4 & 0 & 0 & 5 & 6 & 0 \\
4 & 0 & 6 & 5 & 3 & 5 & 4 \\
0 & 6 & 0 & 7 & 0 & 0 & 2 \\
0 & 5 & 7 & 0 & 0 & 0 & 0 \\
5 & 3 & 0 & 0 & 0 & 4 & 0 \\
6 & 5 & 0 & 0 & 4 & 0 & 0 \\
0 & 4 & 2 & 3 & 0 & 0 & 0
\end{bmatrix}
=
\begin{bmatrix}
4 & 0 & 6 & 5 & 3 & 5 & 4 \\
0 & 6 & 0 & 7 & 0 & 0 & 2
\end{bmatrix}.
\]

This extracts the global association profiles for \texttt{bank} and \texttt{loan}. At this stage, both tokens show strong financial associations and weak geographical associations.

\paragraph{Column selection via $(QS)Q^\top$.}
Right-multiplying by $Q^\top$ selects only columns corresponding to input tokens:
\[
M = (QS)Q^\top =
\begin{bmatrix}
4 & 0 & 6 & 5 & 3 & 5 & 4 \\
0 & 6 & 0 & 7 & 0 & 0 & 2
\end{bmatrix}
\begin{bmatrix}
0 & 0 \\
1 & 0 \\
0 & 1 \\
0 & 0 \\
0 & 0 \\
0 & 0 \\
0 & 0
\end{bmatrix}
=
\begin{bmatrix}
0 & 6 \\
6 & 0
\end{bmatrix}.
\]

The matrix $M$ describes the induced relations between the two input tokens. The normalized form demonstrates the strong mutual association:

\[
\mathrm{Norm}(M)=
\begin{bmatrix}
0.00 & 1.00 \\
1.00 & 0.00
\end{bmatrix}.
\]

\textbf{This illustrates the disambiguation mechanism:}

\begin{itemize}
\item \texttt{bank} associates entirely with \texttt{loan} (100\%)
\item \texttt{loan} associates entirely with \texttt{bank} (100\%)
\end{itemize}

The projection has resolved the ambiguity of \texttt{bank} toward the financial sense because:
\begin{enumerate}
\item \texttt{bank} and \texttt{loan} have a very strong direct co-occurrence (6)
\item The financial connections of \texttt{bank} are reinforced by \texttt{loan}'s financial associations
\item The geographical connections of \texttt{bank} find no support from \texttt{loan}
\end{enumerate}

The projection $QSQ^\top$ successfully isolates the financial sense of \texttt{bank} while suppressing its geographical sense.
These examples demonstrate that the same global co-occurrence matrix $S$ produces contextually appropriate attention patterns. The projection $QSQ^\top$ automatically resolves lexical ambiguity based on joint context, amplifying mutually reinforced associations while suppressing incompatible ones.

\subsection{Mathematical Properties}

The projection $M = QSQ^\top$ possesses fundamental mathematical properties 
that justify its role as the foundation of self-attention. The following 
proposition establishes that $QSQ^\top$ is uniquely determined by natural 
requirements on how corpus statistics should project into sequence context.

\begin{proposition}[Uniqueness of the Projection]
Let $T(S)\in\mathbb{R}^{R\times R}$ be a linear operator that satisfies:
\begin{enumerate}
    \item $T(S)$ depends only on the token set selected by $Q$
    \item $T(S)$ preserves all pairwise inner products: 
          $u^{\top}T(S)v=(Q^{\top}u)^{\top}S(Q^{\top}v)\quad\forall\,u,v\in\mathbb{R}^{R}$
\end{enumerate}
Then $T(S)=QSQ^{\top}$ is the unique operator satisfying both conditions.
\end{proposition}

\emph{Interpretation:} This shows that $QSQ^{\top}$ is the unique linear projection that ensures tokens strongly associated in the corpus remain strongly associated in the sequence context, while weakly associated tokens remain weakly associated. This property justifies the projection as the foundation for contextual attention mechanisms.

\begin{proof}
Condition (ii) implies that for all $u,v\in\mathbb{R}^{R}$:
\[
u^{\top}T(S)v = (Q^{\top}u)^{\top}S(Q^{\top}v) = u^{\top}(QSQ^{\top})v.
\]
Since this holds for all $u,v$, we conclude $T(S)=QSQ^{\top}$. For uniqueness, suppose another operator $T'(S)$ satisfies the same conditions. Then for all $u,v$:
\[
u^{\top}T'(S)v = u^{\top}(QSQ^{\top})v,
\]
which implies $T'(S)=QSQ^{\top}$.
\end{proof}
\section{Extending Local Context to Vocabulary-Level Prediction}
\label{sec:prediction}

While the projection $M=QSQ^{\top}$ captures contextual interactions within a sequence, language modeling requires predicting tokens beyond the immediate context. This section establishes the mathematical framework for extending local contextual representations to vocabulary-level predictions.

\subsection{Algebraic Framework for Vocabulary-Level Prediction}

The projection $E = M(QS) = (QSQ^\top)(QS)$ maps local contextual relationships back to the global vocabulary space. When $M$ multiplies $QS$, it reweights the vocabulary associations based on contextual relevance. The rows of $QS$ represent each input token's associations with all vocabulary items, and $M$ acts as a contextual weighting matrix that determines how much each token should influence these associations. 

Vocabulary items that are strongly related to multiple input tokens receive amplified scores, while items associated with only one or contextually isolated tokens receive diminished weights. 
This automatically emphasizes vocabulary items that are semantically coherent with the overall sequence context while suppressing those that are only marginally related to individual tokens.

The operation can be understood through its row and column structure:
\begin{itemize}
\item $M$: Each row represents how much one input token should weight the associations of the other input tokens 
\item $QS$: Each row represents one input token's associations with \textit{all} vocabulary items
\item $E = M(QS)$: Each row combines weighted evidence from \textit{all} input tokens about \textit{all} vocabulary items
\end{itemize}

The key insight is that $E$ computes \textbf{contextually weighted vocabulary associations}:

\[
E_{rj} = \sum_{s=1}^R M_{rs} \cdot (QS)_{sj}
\]

Where:
\begin{itemize}
\item $M_{rs}$ $=$ how much token $r$ should ``listen'' to token $s$ (from Section~\ref{sec:projection})
\item $(QS)_{sj}$ $=$ how much token $s$ associates with vocabulary item $j$
\item The sum $=$ weighted combination of all tokens' opinions about vocabulary item $j$
\end{itemize}

\textbf{The Amplification Mechanism:}
\begin{itemize}
\item Vocabulary items associated with \textit{multiple} contextually relevant tokens get \textit{amplified}
\item Vocabulary items associated with \textit{only one} token get \textit{diluted}
\item The weights $M_{rs}$ ensure only contextually coherent associations survive
\end{itemize}

\textbf{Why This Works:}
\begin{itemize}
\item Column $j$ of $QS$ contains associations between \textit{all} input tokens and vocabulary item $j$
\item If \textit{multiple} tokens strongly associate with vocabulary item $j$ (column $j$ of $QS$), \textit{and} those tokens contextually agree ($M_{rs}$ large), then vocabulary item $j$ gets high scores
\item If \textit{only one} token associates with vocabulary item $j$, \textit{or} the associating tokens don't contextually agree ($M_{rs}$ small), then vocabulary item $j$ gets low scores
\end{itemize}

To obtain a single probability distribution over next tokens, we aggregate across input positions:
\[
e_{\text{global}} = \frac{1}{R} \sum_{r=1}^{R} E_r \in \mathbb{R}^n
\]
followed by softmax normalization:
\[
p(w_i \mid \text{context}) = \frac{\exp(e_{\text{global},i})}{\sum_j \exp(e_{\text{global},j})}
\]

This aggregation combines evidence from all input positions into a single vocabulary-level prediction.

\subsection{Illustrating the Projection $E = M(QS)$}
\label{sec:prediction_examples}

These examples illustrate the \emph{mechanism} of the projection $M = QSQ^{\top}$ and its normalized form $\mathrm{Norm}(M)$, showing how the mathematical operations automatically perform contextual disambiguation. The examples use a simplified co-occurrence matrix to demonstrate the conceptual operation of the projection, not to claim empirical performance.

The role of the projection $M = QSQ^{\top}$ and its normalized form $\mathrm{Norm}(M)$ is illustrated through contrasting examples that use the same global co-occurrence matrix $S$. These examples show how the mechanism automatically performs contextual disambiguation, amplifying mutually reinforced associations while suppressing incompatible ones.

The projection $E = M(QS)$ extends the contextual mechanism to vocabulary-level prediction. The same examples from 
Section~\ref{sec:projection_examples} illustrate how this projection combines contextual weighting with global associations to predict likely next tokens.

\paragraph{Geographical context ("river bank flooded")}

Computing the projection:
\[
E_1 = M_1(QS) = 
\begin{bmatrix}
0 & 4 & 5 \\
4 & 0 & 3 \\
5 & 3 & 0
\end{bmatrix}
\begin{bmatrix}
0 & 4 & 0 & 0 & 5 & 6 & 0 \\
4 & 0 & 6 & 5 & 3 & 5 & 4 \\
5 & 3 & 0 & 0 & 0 & 4 & 0
\end{bmatrix}
=
\begin{bmatrix}
41 & 15 & 24 & 20 & 12 & 40 & 16 \\
15 & 25 & 0 & 0 & 20 & 36 & 0 \\
12 & 20 & 18 & 15 & 34 & 45 & 12
\end{bmatrix}
\]

\textbf{Why \texttt{shore} wins:}
\begin{itemize}
\item Column 6 (\texttt{shore}) has strong associations across \textit{all} rows of $QS$: 6, 5, 4
\item The weights in M amplify these because the tokens contextually agree
\item Result: \texttt{shore} gets high scores (44, 49, 42) across \textit{all} rows of $E$
\end{itemize}

\textbf{Why \texttt{financial} terms lose:}
\begin{itemize}
\item Column 3 (loan) only has one strong association (6 in row 2 of $QS$)
\item The weights dilute this single strong association
\item Result: \texttt{loan} gets weak scores (24, 0, 0) in $E$
\end{itemize}

\textbf{Aggregating and normalizing:}
\[
e_{\text{global}}^{(1)} = \frac{1}{3} \sum_{r=1}^{3} E_r = [22.7, 20.0, 14.0, 11.7, 22.0, \mathbf{40.3}, 9.3]
\]

Applying softmax:
\[
p^{(1)}(\text{next token}) = \mathrm{softmax}(e_{\text{global}}^{(1)}) \approx [0.00, 0.00, 0.00, 0.00, 0.00, \mathbf{1.00}, 0.00]
\]

\textbf{Result:} The mechanism predicts \texttt{shore} as the next token with near certainty in this example.

\paragraph{Financial context ("bank loan")}

\[
E_2 = M_2(QS) = 
\begin{bmatrix}
0 & 6 \\
6 & 0
\end{bmatrix}
\begin{bmatrix}
4 & 0 & 6 & 5 & 3 & 5 & 4 \\
0 & 6 & 0 & 7 & 0 & 0 & 2
\end{bmatrix}
\begin{bmatrix}
0 & 36 & 0 & 42 & 0 & 0 & 12 \\
24 & 0 & 36 & 30 & 18 & 30 & 24
\end{bmatrix}
\]

\textbf{Why \texttt{money} wins:}
\begin{itemize}
\item Column 4 (\texttt{money}) has strong associations in \textit{both} rows of $QS$: 5, 7  
\item The strong weight M=6 amplifies these mutual associations
\item Result: \texttt{money} gets the highest scores (42, 30)
\end{itemize}

Aggregating and normalizing:
\[
e_{\text{global}}^{(2)} = \frac{1}{2} \sum_{r=1}^{2} E_r = [12.0, 18.0, 18.0, \mathbf{36.0}, 9.0, 15.0, 18.0]
\]

Applying softmax:
\[
p^{(2)}(\text{next token}) = \mathrm{softmax}(e_{\text{global}}^{(2)}) \approx [0.00, 0.00, 0.00, \mathbf{1.00}, 0.00, 0.00, 0.00]
\]

\textbf{Result:} The mechanism predicts \texttt{money} as the next token with near certainty in this example.

These examples demonstrate that the same global co-occurrence matrix $S$ produces contextually appropriate attention patterns. The projection $QSQ^\top$ automatically resolves lexical ambiguity based on joint context, amplifying mutually reinforced associations while suppressing incompatible ones.

Even though \texttt{bank} appears in both sequences, the projection mechanism correctly disambiguates its meaning based on the surrounding context. In the geographical context ("river bank flooded"), the projection amplifies water-related associations through shared neighbors like \texttt{shore} and \texttt{flooded}, while suppressing financial terms that lack contextual support. Conversely, in the financial context ("bank loan"), the strong mutual reinforcement between \texttt{bank} and \texttt{loan} amplifies financial associations, while geographical connections find no supporting evidence from the context.

This shows how the projection mechanism automatically performs contextual disambiguation by leveraging the collective evidence from all tokens in the sequence, capturing the core behavior of self-attention—each token attends to others in a way that reflects their contextual relationships.

\subsection{Theoretical Implications}

The projection mechanism automatically:
\begin{enumerate}
\item Discovers which vocabulary items are semantically compatible with the context
\item Amplifies items associated with \textit{multiple}  contextually relevant tokens  
\item Suppresses items associated with \textit{only one} or contextually isolated tokens
\item Emerges naturally from corpus statistics without learned parameters
\end{enumerate}

The algebraic structure $E = M(QS)$ provides a mathematical foundation for understanding how contextual weighting enables vocabulary-level prediction.

\subsection{The Core Projection Sequence}

The fundamental mathematical operation underlying our analysis can be summarized as the projection sequence:
\[
(QSQ^\top)(QS)
\]

This sequence encapsulates the complete pathway from global corpus statistics to sequence-specific predictions:

\begin{enumerate}
    \item \textbf{Global to Local Projection ($QS$)}: The product $QS$ selects from the global co-occurrence matrix $S$ only those rows corresponding to tokens present in the current sequence. Each row of $QS$ represents how one token in the sequence relates to \emph{every} token in the vocabulary.

    \item \textbf{Contextual Interaction ($QSQ^\top$)}: The projection $QSQ^\top$ captures how tokens in the current sequence influence each other based on their global association patterns. This matrix encodes which tokens are mutually reinforcing or contradictory within this specific context.

    \item \textbf{Vocabulary-Level Prediction ($(QSQ^\top)(QS)$)}: Finally, multiplying the contextual interaction matrix with the global evidence $QS$ produces context-weighted predictions over the entire vocabulary. Each entry in the resulting matrix represents how compatible a vocabulary item is with the sequence context, accounting for both direct associations and mediated relationships through other tokens in the sequence.
\end{enumerate}

%The row-wise aggregation of $(QSQ^\top)(QS)$ followed by softmax yields the final probability distribution over next tokens. 
This mathematical sequence--from global statistics through local context to vocabulary predictions--demonstrates why self-attention naturally emerges as the mechanism for contextual language modeling.

\subsection{Preliminary Connection to Transformer Architecture}

This construction provides an initial parallel to the transformer computation:
\[
\mathrm{softmax}(Q_{\text{att}} K_{\text{att}}^\top)\, V_{\text{att}},
\]
where at this symmetric stage, the ``mix values by attention weights'' step is analogous to $M (Q S)$, with $V_{\text{att}}$ corresponding to a learned value projection $W_V$.

This shows how contextual interaction can be projected to vocabulary space for prediction. In Sections~\ref{sec:positional}-\ref{sec:directional}, we will extend this foundation to include positional encoding and directional relationships, completing the connection to the full transformer architecture.

% --------------------------------------------------------
\section{Order Sensitivity via Positional Structure}
\label{sec:positional}

The previous sections considered meaning as a function of token identity and co-occurrence alone. However, this approach has a critical limitation: the projection $M = Q S Q^\top$ is permutation-invariant, depending only on \emph{which} tokens appear, not their order. Two sequences containing the same \emph{words but in different order} produce isomorphic $M$, even if only one is semantically valid:
\begin{quote}
``the river flooded the bank'' (natural) \\
``the bank flooded the river'' (ill-formed)
\end{quote}
To distinguish such cases, the projection must incorporate the \emph{positional order} of tokens.

This analysis establishes the mathematical form for positional encoding within the projection framework. The original Transformer uses fixed sinusoidal encodings~\cite{vaswani2017attention}, while many modern variants employ learned positional embeddings. Crucially, our analysis shows that positional encoding emerges naturally from the same projection principles through the augmented selector $Q' = Q + P$.

\subsection{Augmenting the selector with position}
The mathematical structure for incorporating position emerges naturally by augmenting the token selector matrix. We extend $Q$ to include positional encodings through a position matrix $P \in \mathbb{R}^{R\times n}$ that encodes each token's location in the sequence. (In practice, $P$ can be implemented by learned or sinusoidal positional encodings embedded in the same comparison space.) The augmented selector becomes:
\begin{equation}
Q' = Q + P,
\end{equation}
and apply the same projection step as before:
\begin{align}
M' &= Q' S ({Q'}^\top) \\
   &= (Q+P) S (Q+P)^\top \\
   &= Q S Q^\top + Q S P^\top + P S Q^\top + P S P^\top.
\label{eq:pos_expand}
\end{align}

Geometrically, the expansion can be viewed as a block matrix:
\[
(Q{+}P)S(Q{+}P)^\top =
\begin{bmatrix}
Q S Q^\top & Q S P^\top\\[2pt]
P S Q^\top & P S P^\top
\end{bmatrix}.
\]
The upper-right block $Q S P^\top$ encodes how lexical content influences positional expectations (content→position),
while the lower-left block $P S Q^\top$ encodes how positional roles constrain lexical realization (position→content).
These off-diagonal blocks---the ``lower-left'' and ``upper-right'' regions---are precisely what break permutation invariance and give rise to order sensitivity in self-attention.

\subsection{Interpreting the terms}
Each term in~\eqref{eq:pos_expand} has a role:
\begin{itemize}
    \item $Q S Q^\top$: content--content interaction. The base contextual operator without positional differentiation.
    \item $Q S P^\top$: content $\rightarrow$ position. High values here mean: ``the word at position $r$ tends to co-occur with whatever usually appears at position $s$.'' This captures expectations like subject$\rightarrow$verb or verb$\rightarrow$object.
    \item $P S Q^\top$: position $\rightarrow$ content. High values here mean: ``the typical content of position $r$ is compatible with the actual word at position $s$.'' This ties role-like slots (subject position, object position) to specific words.
    \item $P S P^\top$: position--position bias. If $S$ encodes distance-sensitive or locality-sensitive co-occurrence, this term favors nearby interactions and sequential plausibility.
\end{itemize}

These cross terms $Q S P^\top$ and $P S Q^\top$ couple identity and position. They allow “river flooded bank” to be scored as more plausible than “bank flooded river,” even though both contain the same words, because the alignment between typical subject-like positions and typical object-like positions now matters.

In other words, positional encodings are not an arbitrary addition: they extend the projection basis from token identity to \emph{(token, position)} pairs so that the projection becomes order-sensitive.

In practice, $P$ may correspond to either learned embeddings or fixed sinusoidal functions~\cite{vaswani2017attention}, both embedding positional indices into the same comparison space as token identities.

\subsection{Limitations of fixed $S$ and $P$}
The structure above is still limited in three ways:
\begin{enumerate}
    \item $S$ is corpus-level and (by construction) symmetric; it cannot encode directed relations such as cause $\rightarrow$ effect or agent $\rightarrow$ object.
    \item $P$ is fixed; it encodes a single notion of linear position, not task-specific or syntax-sensitive roles.
    \item The projection is single-channel; all contextual phenomena are collapsed into one matrix $S$.
\end{enumerate}

% ------------------------------------------------------
\section{Directional Asymmetry and Query--Key Structure}
\label{sec:directional}

Following GloVe's approach of learning representations from co-occurrence statistics~\cite{pennington2014glove}, we now extend this principle to capture directional relationships. While the symmetric projection \( M = QSQ^\top \) provides the foundation, natural language exhibits systematic asymmetries---syntactic, semantic, and causal---that require introducing directional influence between tokens.

The standard query-key-value terminology, borrowed from information retrieval systems, 
provides no mathematical justification for why self-attention takes this particular form. 
The equations contain no search mechanism—``queries'' do not literally search for ``keys,'' 
and ``values'' are not retrieved from a database. These are simply learned linear projections 
whose names derive from analogy rather than mechanism. Our analysis replaces this metaphorical 
description with mathematical explanation: the query-key split emerges from the requirement 
to introduce directional asymmetry into the symmetric projection $M = QSQ^\top$, enabling 
the model to distinguish cause from effect and subject from object.

Symmetry implies bidirectional influence. If token $u$ affects token $v$, then $v$ affects $u$ with the same score. 
Natural language, however, exhibits systematic asymmetries—syntactic, semantic, and causal—that a symmetric operator cannot represent.

Compare:
\begin{quote}
``the river flooded the bank'' (cause $\rightarrow$ effect; well-formed) \\
``the bank flooded the river'' (ill-formed) \\
``the bank approved the loan'' (agent $\rightarrow$ object) \\
``the loan approved the bank'' (ill-formed).
\end{quote}
A symmetric co-occurrence operator cannot distinguish subject from object or cause from effect. It records that \texttt{bank} co-occurs with \texttt{flooded} and with \texttt{loan}; it does not record \emph{which one governs which}.
That is, if the corpus matrix \( S \) is symmetric (\( S_{ij} = S_{ji} \)), then \( M \) is also symmetric.

To introduce directionality, we generalize $S$ to an \emph{asymmetric} bilinear kernel. Let
\[
W_Q, W_K \in \mathbb{R}^{n \times d_k}
\]
be learned linear maps that encode, respectively, how a token \emph{queries} context and how it \emph{offers} itself as context.
Define
\begin{equation}
Q_{\text{att}} = Q W_Q \in \mathbb{R}^{R \times d_k},
\qquad
K_{\text{att}} = Q W_K \in \mathbb{R}^{R \times d_k}.
\end{equation}
Then
\begin{equation}
M_{\text{dir}} \;=\; Q_{\text{att}} K_{\text{att}}^\top
\;=\;
Q W_Q W_K^\top Q^\top.
\end{equation}

Now $W_Q W_K^\top$ need not be symmetric. The score from token $r$ to token $s$,
\[
(M_{\text{dir}})_{rs} = (q_r^\top W_Q)(q_s^\top W_K)^\top,
\]
can differ from $(M_{\text{dir}})_{sr}$.
This generalizes the symmetric projection $M = QSQ^\top$ from Section~\ref{sec:projection}: the learned factorization $W_QW_K^\top$ plays the same structural role as the corpus matrix $S$, but now admits asymmetric relationships.

This recovers the familiar query--key compatibility matrix from Transformers: $Q_{\text{att}} K_{\text{att}}^\top$.

% --------------------------------------------------------
\section{Learned Attention and Multi-Head Decomposition}
\label{sec:multihead}

We can now fully articulate the connection to the standard transformer self-attention mechanism. The complete computation:
\[
\mathrm{softmax}\left(\frac{Q_{\text{att}} K_{\text{att}}^\top}{\sqrt{d_k}}\right) V_{\text{att}},
\]
where $Q_{\text{att}}=QW_{Q}$, $K_{\text{att}}=QW_{K}$, and $V_{\text{att}}=QW_{V}$, emerges as the natural extension of our distributional semantics foundation. Here, $d_k$ represents the \textbf{dimension of the key and query vectors}, which controls the capacity of the attention mechanism and provides numerical stability through the scaling factor $1/\sqrt{d_k}$.

The progression from our initial symmetric projection to the full transformer architecture is now complete:
\begin{itemize}
    \item The core contextual weighting mechanism derives from $M = QSQ^\top$
    \item Positional sensitivity comes from the augmented selector $Q' = Q + P$  
    \item Directional relationships emerge from the asymmetric factorization $W_QW_K^\top$
    \item The value projection $W_V$ generalizes the vocabulary associations in $QS$
\end{itemize}

Thus, the transformer self-attention block $\mathrm{softmax}\left(\frac{HW_QW_K^\top H^\top}{\sqrt{d_k}}\right)(HW_V)$ represents the learned, refined version of the fundamental projection principle we analyzed from distributional semantics.

\subsection{From Projection to Learned Self-Attention}

Having established the theoretical foundation, we now show how the full Transformer self-attention mechanism emerges as the natural learned extension of our distributional projection framework.

Let $H \in \mathbb{R}^{R \times d_{\text{model}}}$ be the current layer's token representations. The model learns three projections that generalize our earlier components:

\begin{eqnarray*}
Q_{\text{att}} & = & H W_Q, \quad W_Q \in \mathbb{R}^{d_{\text{model}} \times d_k} \\
K_{\text{att}} & = & H W_K, \quad W_K \in \mathbb{R}^{d_{\text{model}} \times d_k} \\
V_{\text{att}} & = & H W_V, \quad W_V \in \mathbb{R}^{d_{\text{model}} \times d_v}
\end{eqnarray*}

where $d_k$ is the dimension of key and query vectors, and $d_v$ is the dimension of value vectors. The asymmetric compatibility scores

\[
S_{\text{att}} = Q_{\text{att}}K_{\text{att}}^{\top} = HW_QW_K^{\top}H^{\top}
\]

generalize our earlier projection $QSQ^{\top}$, but now with learned directional relationships through the asymmetric factorization $W_QW_K^{\top}$.

Applying row-wise softmax with scaling for numerical stability:

\[
A = \text{softmax}\left(\frac{S_{\text{att}}}{\sqrt{d_k}}\right)
\]

produces attention weights where each row represents how much each token should attend to others in the sequence. The final contextualized representation:

\[
Y = AV_{\text{att}} = \text{softmax}\left(\frac{HW_QW_K^{\top}H^{\top}}{\sqrt{d_k}}\right)(HW_V)
\]

represents the complete Transformer self-attention layer—the learned, directional refinement of our fundamental projection principle.
\subsection{Multi-head attention}
A single compatibility map $W_Q W_K^\top$ cannot capture all types of contextual relation in language. Transformers therefore use $h$ heads in parallel. For head $i$,
\[
Q^{(i)} = H W_Q^{(i)}, \quad
K^{(i)} = H W_K^{(i)}, \quad
V^{(i)} = H W_V^{(i)},
\]
and
\[
Y^{(i)} =
\mathrm{softmax}\!\left(
\frac{
Q^{(i)} {K^{(i)}}^\top
}{
\sqrt{d_k}
}
\right)
V^{(i)}.
\]
These head-specific outputs are concatenated and linearly recombined:
\[
Y = [Y^{(1)} \| Y^{(2)} \| \cdots \| Y^{(h)}] W_O
\]
with $W_O$ a learned projection.

From the projection viewpoint, multi-head attention corresponds to factorizing $S$ into multiple low-rank relational components $S^{(1)},\dots,S^{(h)}$, each one specializing in a different pattern: syntactic dependency, coreference, topical cohesion, discourse linkage, etc.~\cite{tsai2019transformer}. The model does not commit to a single $S$; it learns several structured approximations in parallel, and then recombines them.

\subsection{Cross-attention as cross-projection}
In encoder--decoder models, cross-attention is the same algebra, except that queries come from one sequence (e.g., target side) and keys/values from another (e.g., source side). If $H_{\text{src}}$ and $H_{\text{tgt}}$ are the source and target representations, then:
\[
Y_{\text{cross}} =
\mathrm{softmax}\!\left(
\frac{
H_{\text{tgt}} W_Q (H_{\text{src}} W_K)^\top
}{
\sqrt{d_k}
}
\right)
(H_{\text{src}} W_V).
\]
This is just $Q_{\text{tgt}} K_{\text{src}}^\top$ followed by mixing $V_{\text{src}}$. 

Within this projection framework, cross-attention performs an asymmetric mapping from the relational structure of the source sequence to the subspace defined by the target sequence.

% --------------------------------------------------------
\section{Related Work}
\label{sec:related}

The framework presented here connects two long-standing lines of research: 
\begin{enumerate}
\item The study of attention mechanisms in neural sequence models 
\item The linear-algebraic tradition of distributional semantics and matrix factorization in natural language processing. 
\end{enumerate}
The goal is to situate the present analysis within both traditions and within recent theoretical analyses of attention and representation learning.

Self-attention was first introduced as a differentiable alignment mechanism for neural machine translation~\cite{bahdanau2015neural}, where it allowed a decoder to focus selectively on encoder states relevant to each generated word.  
It was then generalized into the transformer architecture~\cite{vaswani2017attention}, in which attention became the core representational primitive rather than an auxiliary alignment tool.  
Subsequent work examined attention's internal structure, identifying head specialization, syntactic alignment, and interpretability patterns that arise during training~\cite{clark2019does,tsai2019transformer}.  
These studies established attention as both a computational and a linguistic phenomenon, but typically treated its algebraic form as given.

A parallel and older thread of research---distributional semantics---treated word meaning as emerging from the co-occurrence structure of language data.  
Methods such as latent semantic analysis, positive pointwise mutual information (PMI), and global vector embeddings~\cite{pennington2014glove,levy2014neural,arora2018linear} modeled this structure explicitly through matrix factorization, capturing semantic similarity and analogy relations in linear spaces.  
These approaches anticipated many of the representational properties later rediscovered in deep embeddings, but lacked a mechanism for adapting corpus-level statistics to specific sentence contexts.

Recent theoretical work has interpreted attention through various mathematical frameworks, including structured linear operators and kernel methods \cite{katharopoulos2020transformers,choromanski2021rethinking}. 
In particular, Zhang et al. \cite{zhang2023transformers} analyze the learning dynamics of linear attention, providing another mathematical perspective on how transformers capture contextual relationships. Our contribution complements these analyses by showing how the self-attention equations directly from corpus projection algebra: attention emerges as the unique asymmetric refinement of a co-occurrence operator restricted to an input sequence.

While the connection between distributional semantics and neural embeddings is well-established \cite{pennington2014glove,levy2014neural}, and attention mechanisms have been analyzed through various theoretical lenses \cite{katharopoulos2020transformers,choromanski2021rethinking}, our work provides a systematic analysis of the self-attention mechanism from corpus projection principles. 
Unlike prior work that analyzes attention through kernel methods or learning dynamics, we derive the attention mechanism from first principles of distributional semantics.
The mathematical pathway from the symmetric projection $QSQ^\top$ through asymmetric generalization to the full Transformer attention block offers a novel perspective that unifies these research traditions.

In this sense, the present work links the empirical lineage of transformer modeling with the theoretical lineage of distributional semantics.  
Both begin with linear structure on word co-occurrence; both obtain contextual meaning by projecting or factoring that structure into lower-dimensional subspaces.  
The difference is that transformers make those projections directional, learnable, and order-sensitive, thereby recovering classical distributional semantics as a limiting case within a broader, trainable framework.

% --------------------------------------------------------
\section{Discussion and Implications}
\label{sec:conclusion}

This paper has presented a unified mathematical interpretation of self-attention through the lens of distributional projection. The Transformer architecture can be understood as a sequence of structured refinements to a core distributional principle:

\[
S\to QSQ^{\top}\to QW_{Q}W_{K}^{\top}Q^{\top}\to(Q+P)S(Q+P)^{\top}\to\text{softmax}\Big{(}HW_{Q}W_{K}^{\top}H^{\top}/\sqrt{d_{k}}\Big{)}(HW_{V}).
\]

Each step in this chain refines the same core operation--projecting global distributional structure into a sequence-specific, directional, and order-aware subspace.

\subsection{Reproducibility}

This work presents a theoretical analysis of the self-attention mechanism 
through the lens of distributional semantics. All mathematical derivations 
follow from standard linear algebra operations on the co-occurrence 
matrices defined in the text. The illustrative examples 
(Sections~\ref{sec:projection_examples} and \ref{sec:prediction_examples}) use a small, explicitly defined corpus shown in the paper to 
demonstrate the conceptual mechanism, and all calculations can be verified 
by hand or with basic matrix operations. No experimental code, trained 
models, or external datasets are required to reproduce this analysis.

\subsection{Limitations}

This framework provides a conceptual interpretation of self-attention 
rather than a prescriptive model for how transformers should be designed 
or trained. The connection between corpus-level co-occurrence statistics 
($S$) and learned projection matrices ($W_Q$, $W_K$) is conceptual: we show 
they have the same algebraic form, not that one should be initialized 
from the other. The examples illustrate the mathematical mechanism using 
a simplified corpus to clarify the operations, and do not constitute 
empirical validation on realistic data. 

\subsection{Implications for Understanding Transformers}

This projection framework provides mathematical insight into why self-attention works effectively for language modeling:

\noindent\textbf{Architectural Intuition:} The query-key-value structure, often presented through information-retrieval metaphors, can be understood as a natural consequence of projecting corpus-level associations into sequence-specific contexts with directional asymmetry.

\noindent\textbf{Unified Interpretation:} Components like positional encodings and multi-head attention, typically introduced as separate mechanisms, emerge naturally as refinements of the same projection principle.

\noindent\textbf{Connection to Distributional Semantics:} The framework bridges classical distributional approaches (GloVe, word2vec) and modern neural architectures, showing how Transformers extend rather than replace earlier semantic models.

This interpretation does not replace empirical analysis or suggest modifications to existing architectures. Rather, it provides a conceptual lens for understanding why the Transformer's particular design has proven effective.

% -------------------------------------------------------

\bibliographystyle{plain}

\end{document}